\pdfoutput=1 
\documentclass[letterpaper, 10 pt, conference]{ieeeconf}  

\IEEEoverridecommandlockouts                              

\usepackage{amsmath} 
\usepackage{amssymb}  
\usepackage{hyperref}
\usepackage{gensymb}
\usepackage{comment}
\usepackage{mathtools}
\usepackage{flushend} 
\usepackage{tikz} 
\usepackage{float}

\usepackage{caption}
\usepackage{subcaption}

\usepackage{tikz}
\usetikzlibrary{shapes.geometric, arrows, positioning}
\tikzstyle{startstop} = [rectangle, rounded corners, minimum width=2.5cm, minimum height=1.1cm,text centered, draw=black, fill=red!30]
\tikzstyle{io} = [trapezium, trapezium left angle=70, trapezium right angle=110, minimum width=2.5cm, minimum height=1.1cm, text centered, draw=black, fill=blue!30]
\tikzstyle{process} = [rectangle, minimum width=2.5cm, minimum height=1.1cm, text centered, draw=black, fill=orange!30]
\tikzstyle{decision} = [diamond, minimum width=2.5cm, minimum height=1.1cm, text centered, draw=black, fill=green!30]
\tikzstyle{arrow} = [thick,->,>=stealth]

\usepackage{url}
\hypersetup{hidelinks}

\title{\LARGE \bf
Two algorithms for vehicular obstacle detection in sparse pointcloud
}

\author{ Simone Mentasti$^{1}$, Matteo Matteucci$^{1}$, Stefano Arrigoni$^{2}$, Federico Cheli$^{2}$ 
\thanks{Partially supported by project TEINVEIN: TEcnologie INnovative per i VEicoli Intelligenti, CUP (Codice Unico Progetto - Unique Project Code): E96D17000110009 - Call "Accordi per la Ricerca e l'Innovazione", cofunded by POR FESR 2014-2020 (Programma Operativo Regionale, Fondo Europeo di Sviluppo Regionale – Regional Operational Programme, European Regional Development Fund).}
\thanks{$^{1}$ S. Mentasti, M. Matteucci are with the Department of Electronics Information and Bioengineering of Politecnico di Milano, p.zza Leonardo da Vinci 32, Milan, Italy, {\tt\small name.surname@polimi.it}}%
\thanks{$^{2}$ S. Arrigoni, F. Cheli are with the Department of Mechanical Engineering of Politecnico di Milano, via La Masa 1, Milan, Italy, {\tt\small name.surname@polimi.it}}%
}

\begin{document}
\maketitle
\thispagestyle{empty}
\pagestyle{empty}
%
%
\begin{abstract}
One of the main components of an autonomous vehicle is the obstacle detection pipeline. Most prototypes, both from research and industry, rely on lidars for this task. Pointcloud information from lidar is usually combined with data from cameras and radars, but the backbone of the architecture is mainly based on 3D bounding boxes computed from lidar data. To retrieve an accurate representation, sensors with many planes, e.g., greater than 32 planes, are usually employed. The returned pointcloud is indeed dense and well defined, but high-resolution sensors are still expensive and often require powerful GPUs to be processed. Lidars with fewer planes are cheaper, but the returned data are not dense enough to be processed with state of the art deep learning approaches to retrieve 3D bounding boxes. In this paper, we propose two solutions based on occupancy grid and geometric refinement to retrieve a list of 3D bounding boxes employing lidar with a low number of planes (i.e., 16 and 8 planes). Our solutions have been validated on a custom acquired dataset with accurate ground truth to prove its feasibility and accuracy.
\end{abstract}

\section{introduction}

To properly drive in an unknown environment, autonomous vehicles need to sense the surroundings and find possible obstacles. Among the most used sensors for this task are lidars~\cite{ASVADI2016299},~\cite{6083105}~\cite{6122515}. Those sensors return a 3D pointcloud representing the area around the autonomous car up to 100 meters with millimeter precision. The density of this pointcloud is a function of the number of lasers mounted on the sensor, which defines how many planes, sometimes called channels, are in the 3D pointcloud. To retrieve a detailed representation, most autonomous vehicle prototypes employ high-end lidars, generally, with 32 planes or more~\cite{8340798},~\cite{VANBRUMMELEN2018384}. Those sensors provide dense pointclouds that deep learning architectures can process to retrieve 3D bounding boxes of obstacles~\cite{lang2019pointpillars}. However, those sensors are still expensive, and smaller projects can not sustain their costs. Moreover, to process a rich pointcloud in real-time via deep-learning techniques at a high enough frequency to safely control the vehicle, powerful GPUs are required. 

A design-to-cost project might decide to employ sensors with a reduced number of planes, where the returned pointcloud is considerably less defined. For a 16 plane lidar, a human can still identify and classify obstacles, but with a lower resolution, this task becomes challenging. Due to the reduced number of points, deep learning approaches struggle to extract enough features to classify obstacles correctly. For this reason, different methods, not deep-learning-based, need to be employed. 

Lidars and laser sensors have been used in the robotics field way before the rise of autonomous vehicles. Indoor and logistic robots are usually less expensive, and for this reason, they have always been using smaller lidars, usually even single-plane ones. Despite using similar sensor suits, autonomous vehicles and logistics robots have some significant differences.  Indoor robots can navigate the environment with low-level representations such as grid-map~\cite{fankhauser2016universal}; contrarily, planning algorithms for autonomous vehicles, which move at high speed in dynamic environments, require a more high-level representation, like a list of 3D bounding boxes~\cite{bersani2021integrated}. Because of these differences, it is impossible to directly employ classical robotics solutions, which work well with limited plane lidars, on autonomous vehicles. 

\begin{figure}[!t]
        \includegraphics[width=0.5\textwidth]{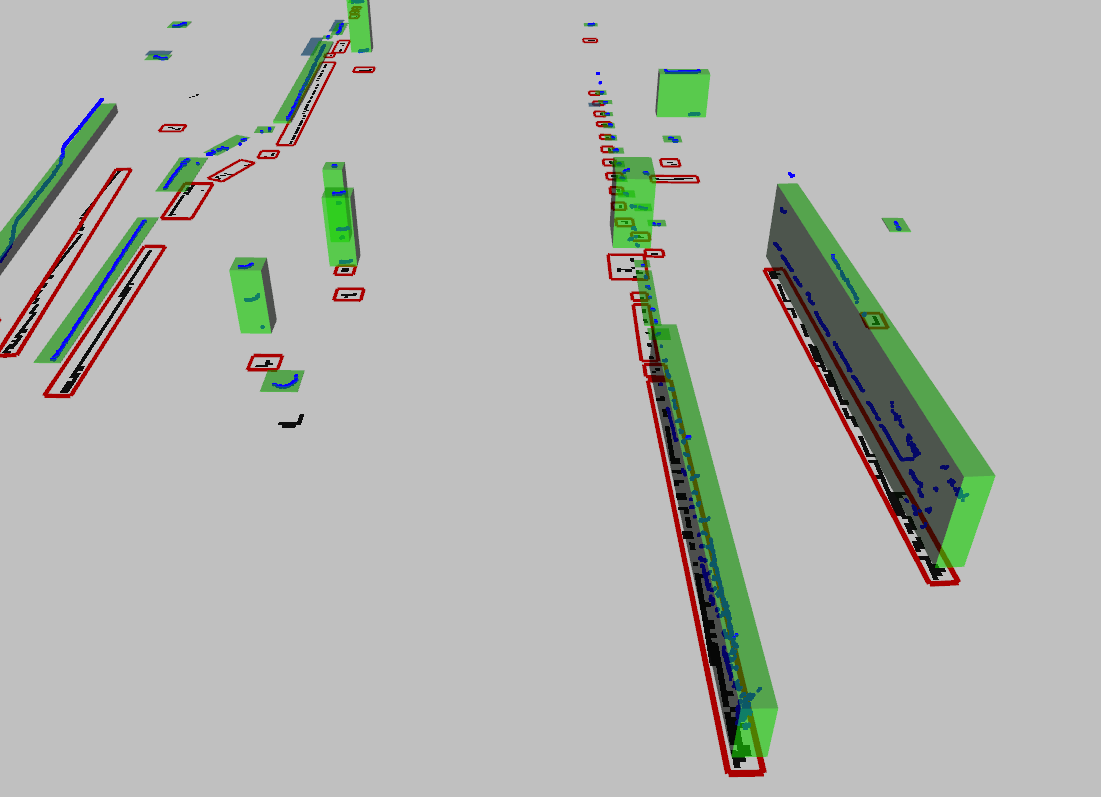}
        \centering
        \caption{Snapshot of the 8 plane obstacle detection algorithm. In blue the original pointcloud points, in black the occupancy grid, and in red and green respectively the 2D bounding boxes on the ground plane and 3D bounding boxes fitted on the pointcloud data.}
        \label{fig:sys}
\end{figure}

In this paper, we propose two solutions, one for a 16 plane lidar and one for an 8 plane lidar. Both approaches are based on occupancy grid, and successive geometric operation on the pointcloud to extract 3D orientated bounding boxes from this low-density pointcloud, as shown in Fig.~\ref{fig:sys}. In particular, the second scenario performs most of the tasks on a 2D grid. Therefore, the proposed solution can be easily extended to any number of plane scenarios (e.g., smaller sensors with four or fewer planes but also higher resolution ones). 

To validate both algorithms, we recorded a custom dataset (publicly available at~\footnote{Dataset will be released upon paper acceptance}), since no dataset is currently available for this task. We employed an RTK-GPS to acquire the exact position of a dynamic obstacle moving around the ego vehicle, equipped with a 16 plane Velodyne. In such a way, we were able to compute the exact position and heading for that specific obstacle to validate our results properly. For the second task, we also performed a decimation of the pointcloud, removing half of the planes to simulate a lower resolution sensor. 

This paper is structured as follows; in Section II, we provide an overview of the current state of the art regarding lidar-based obstacle detection, highlighting the similitude with the robotics world. In Section III and Section IV we describe the two proposed algorithms, for 16 planes lidars and for sensors with fewer planes. Finally, in Section V, we provide experimental validation of both solutions, using our custom recorded dataset, to compare the algorithm output with trustable ground truth. 

\section{related work}

Obstacle detection from laser and lidar data has been a research topic for many years. First solutions come from the indoor robotics world~\cite{44033},\cite{grisettiimproved}; in particular, the most common approach is still to employ a 2D occupancy grid to represent the surrounding of the robot~\cite{30720}. To better generalize on different terrains, and thanks to the increasing computational power and availability of multi-channel sensors, robotics has also moved from 2D grid to 3D representation~\cite{hornung13auro}. 

Autonomous vehicle research started from classical robotics ideas, but rapidly identified different solutions. Nowadays, autonomous cars' detection systems can be divided into three categories: projection methods, volumetric convolutional, and raw pointclouds. The first two are natural evolutions of the robotics approach; projection methods perform a projection of the 3D pointcloud on a 2D plane and then proceed to identify obstacles on the 2D grid. Two class of solution has been identified for this processing phase; the first one is based on geometry and computer vision~\cite{5602377},~\cite{bersani2021integrated}. While the second one leverages on the increased available computational power, employing deep learning techniques to process the 2D grid with convolutional neural network~\cite{BADUE2021113816},~\cite{yang2020lidar}. 

Volumetric convolutional methods are based on a 3D occupancy grid; in this case, the processing phase is similar to the 2D scenario. The 3D voxel grid can again be processed with a convolutional neural network to identify obstacles~\cite{engelcke2017vote3deep},~\cite{zhou2018voxelnet}. The main disadvantage of those approaches is that the 3D grid and its processing are both resource-demanding and require high computational power to be performed in real-time. Nevertheless, they can outperform the 2D approaches thanks to a more accurate representation that retains most of the original pointcloud information.

Raw pointcloud methods perform detection directly on the pointcloud, without processing it. The solutions have rapidly grown in popularity in the last years, thanks to the increasing available computational power, and are one of the most used approaches nowadays. Many solutions are based on the popular architecture PointNet~\cite{qi2017pointnet} and its evolution PointNet++~\cite{qi2017pointnet}, but recently, with the release of big benchmarking dataset, other approaches have emerged~\cite{s18103337}. Deep learning solutions can be extremely powerful and accurate, but they require dense pointclouds, and lidars with a high number of planes to extract enough features. As a result, they perform poorly with excessively sparse pointclouds. 

This paper addresses the scenario where data from lidar are limited in resolution, and obstacles are described only by a few planes. Our approaches are based on the projection method, but we extend it to employ, in the final steps, the original pointcloud data. In such a way, we can achieve higher accuracy in the obstacle pose and heading estimation.

\section{16 plane obstacle detection}

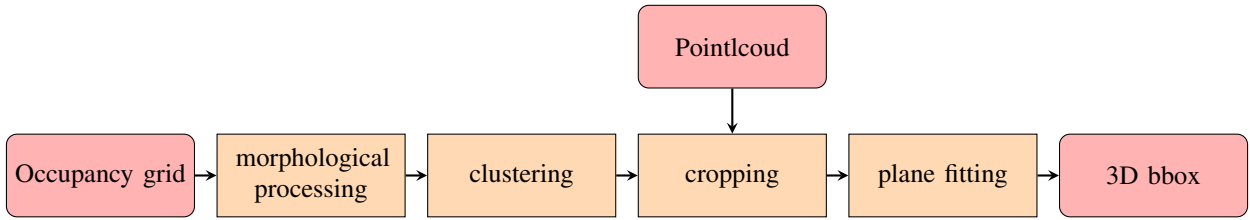
\begin{figure*}[h]
\centering
\begin{tikzpicture}[node distance = 2.8cm and 0cm]

\node (start) [startstop] {Occupancy grid};
\node (p1) [process, right of=start,text width=2cm] {morphological processing};
\node (p2) [process, right of=p1] {clustering};
\node (p3) [process, right of=p2] {cropping};
\node (start2) [startstop, above=of p3, yshift=-2.2cm] {Pointlcoud};
\node (p4) [process, right of=p3] {plane fitting};
\node (stop) [startstop, right of=p4] {3D bbox};
\draw [arrow] (start) -- (p1);
\draw [arrow] (p1) -- (p2);
\draw [arrow] (p2) -- (p3);
\draw [arrow] (start2) -- (p3);
\draw [arrow] (p3) -- (p4);
\draw [arrow] (p4) -- (stop);
\end{tikzpicture}
\caption{Schema of the obstacle detection architecture for 16 plane lidar.}
\label{fig:16_schema}
\end{figure*}

Pointcloud acquired by a 16 plane lidar are generally too sparse to be processed with deep-learning-based solutions to retrieve obstacles. For this reason, the majority of currently developed autonomous vehicles employ at least a 32 planes lidar, or a 64 planes one.
Nevertheless, due to their competitive costs, smaller lidar might be employed in low-cost vehicles or scenarios where computational power is limited, and it is impossible to employ complex neural networks to extract obstacles from the pointcloud. Sixteen plane lidars still return a pointcloud with well-defined obstacles, and therefore it can be processed with geometric-based approaches. In this section, we analyze a geometric solution to obstacle detection using a 16 plane lidar. 

\subsection{Processing pipeline}
The first steps of the pipeline consist of pointcloud preprocessing. Deep learning approaches require the complete pointcloud as input;  in this scenario, instead, it is important to remove all the points that are certainly not obstacles. The first step consists of ground plane removal. This is performed using a slightly modified version of the algorithm proposed in~\cite{zermas2017fast}, to work with our 16 plane lidar. The output of this phase is a pointcloud without the ground plane and the information about the normal of the removed ground. In the next step, we filter all points which belong to objects we are not interested in. In particular, we remove all points above a certain height. This is particularly important to avoid considering as obstacles bridges or light signs. We finally remove points that are too far away on the left and right of the vehicle and therefore are not of interest. The output of this first part is still a 3D pointcloud, but with considerably fewer points, and most importantly, mainly elements that belong to possible obstacles.

In the next step, we proceed to project the 3D pointcloud to a 2D plane, using the normal previously computed when removing the ground plane, and apply a 2D grid on the pointcloud. Iterating through each cell of the grid, we set a cell as `occupied' if the number of projected points falling in that cell is above a threshold. The output of this phase is a 2D occupancy grid, similar to the one employed in mobile robotics and presented in~\cite{mentasti2019multi}. This type of representation is well suited for small and indoor tasks, like logistic robots, but to perform planning in a dynamic environment with an autonomous vehicle it is required a more accurate representation of the surrounding. In particular, the control algorithm performs better with a list of fully characterized obstacles. To retrieve this information from a 2D grid, a further elaboration step is required; the core of our solution is indeed this second processing phase, depicted in Fig.~\ref{fig:16_schema}.
 
The first processing step performed on the occupancy grid is a set of morphological operations. The grid can be easily considered as a binary image. Therefore, classical morphological operations, like closure and opening, can remove single noisy points and merge close areas.
Next, we apply a clustering algorithm on the grid to retrieve a list of connected components. Those elements are all the candidate obstacles. Finally, we iterate through the list, analyzing each cluster; in particular, based on the area and shape of the obstacle, it is possible to remove objects which are not of interest. 

\begin{figure}[t]
     \centering
     \begin{subfigure}[b]{0.3\textwidth}
         \centering
         \includegraphics[width=\textwidth]{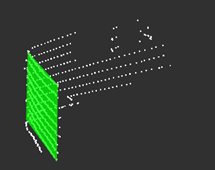}
         \caption{Green plane fitted on the back of the obstacle.}
         \label{fig:plane}
     \end{subfigure}
     \hfill
     \\
     \begin{subfigure}[b]{0.3\textwidth}
         \centering
         \includegraphics[width=\textwidth]{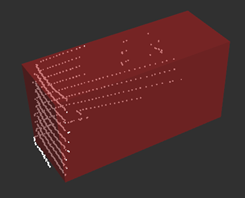}
         \caption{3D bounding box built using the heading from the plane and the size from the pointcloud.}
         \label{fig:box}
     \end{subfigure}

        \caption{Core steps of the 3D bounding box computation process, first a plane is fitted on the back or the side of the obstacle, then a 3D box is computed using the plane heading and the pointcloud data.}
        \label{fig:16_plane_fit}
\end{figure}

In the analyzed scenario, we focus only on finding other vehicles, therefore, all obstacles that are considerably smaller or bigger are removed. Similarly, it is possible to filter only pedestrians based on their shape on the grid.
In the last block of the pipeline, we retrieve a better representation of the obstacles. From the occupancy grid processing phase, it is possible to infer the position and size of the obstacles, but with the discretization of the grid (i.e., the size of a cell). To compute a more accurate representation, we take the list of clusters on the 2D grid and the original pointcloud, and we proceed to crop the small area on the pointcloud relative to each obstacle. Since, in our scenario, we are looking for other vehicles, and the obstacle heading is useful information for the control algorithm, we compute it from the 3D points. In particular, given the pointcloud, we look for a plane perpendicular to the direction of the ego-vehicle, since it will be the side with the highest number of points, as shown in Fig.~\ref{fig:plane}. From the orientation of the plane, it is possible to retrieve the heading of the vehicle. Next, using this plane, we compute the 3D bounding box that best fits all the points, keeping such plane as one of the faces, Fig.~\ref{fig:box}. The final output will be a 3D orientated bounding box that matches the original pointcloud and, therefore, does not have a discretization error. 
\section{8 plane obstacle detection}

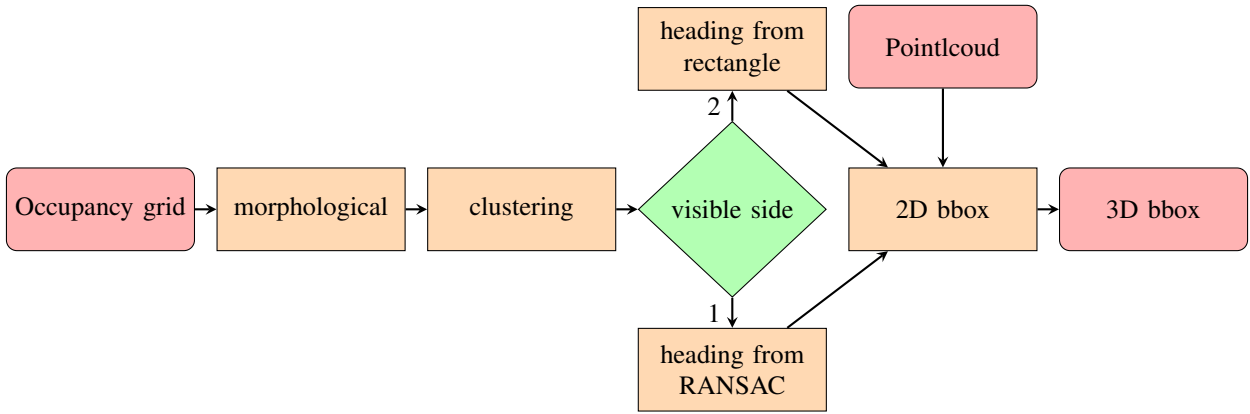
\begin{figure*}[!t]
\centering
\begin{tikzpicture}[node distance = 2.8cm and 0cm]

\node (start_2) [startstop] {Occupancy grid};
\node (p1_2) [process, right of=start_2] {morphological};
\node (p2_2) [process, right of=p1_2] {clustering};
\node (d_2) [decision, right of=p2_2] {visible side};
\node (p3_2) [process, above=of d_2 , yshift=-2.4cm,text width=2cm] {heading from rectangle};
\node (p4_2) [process, below=of d_2 , yshift=2.4cm,text width=2cm] {heading from RANSAC};
\node (p5_2) [process, right of=d_2] {2D bbox};
\node (start2_2) [startstop, above=of p5_2, yshift=-1.75cm] {Pointlcoud};
\node (stop_2) [startstop, right of=p5_2] {3D bbox};
\draw [arrow] (start_2) -- (p1_2);
\draw [arrow] (p1_2) -- (p2_2);
\draw [arrow] (p2_2) -- (d_2);
\draw [arrow] (d_2) --  node [text width=0.2cm,midway,left]{2}(p3_2);
\draw [arrow] (d_2) -- node [text width=0.2cm,midway,left]{1}(p4_2);
\draw [arrow] (p4_2) -- (p5_2);
\draw [arrow] (p3_2) -- (p5_2);
\draw [arrow] (start2_2) -- (p5_2);
\draw [arrow] (p5_2) -- (stop_2);
\end{tikzpicture}
\caption{Schema of the obstacle detection architecture for 8 or less plane lidar.}
\label{fig:8_schema}
\end{figure*}

While 16 plane lidars still provide a dense enough pointcloud to identify obstacles and fit planes to compute 3D bounding boxes, lower density lidars are not suited for this approach. In particular, sensors with a lower number of lasers might provide information of an obstacle only on one or two levels, making it challenging to identify the plane, as for the method we have shown previously. Nevertheless, those sensors are still largely employed, both in lower costs automotive environments and smaller outdoor delivery robots. Therefore the problem of retrieving a list of obstacles from those sensors is still actual. 

\subsection{Processing Pipeline}
Since the previously described approach can not be fully applied with this type of data, the pipeline has a different structure. The first steps are still similar, as shown in Fig~\ref{fig:8_schema}, until we compute the 2D occupancy grid. The only difference will be the size of the cell, which is smaller (i,e,. 0.05 m) to allow us to perform operations directly on the grid. 

\begin{figure}[t]
     \centering
     \begin{subfigure}[b]{0.4\textwidth}
         \centering
         \includegraphics[width=\textwidth]{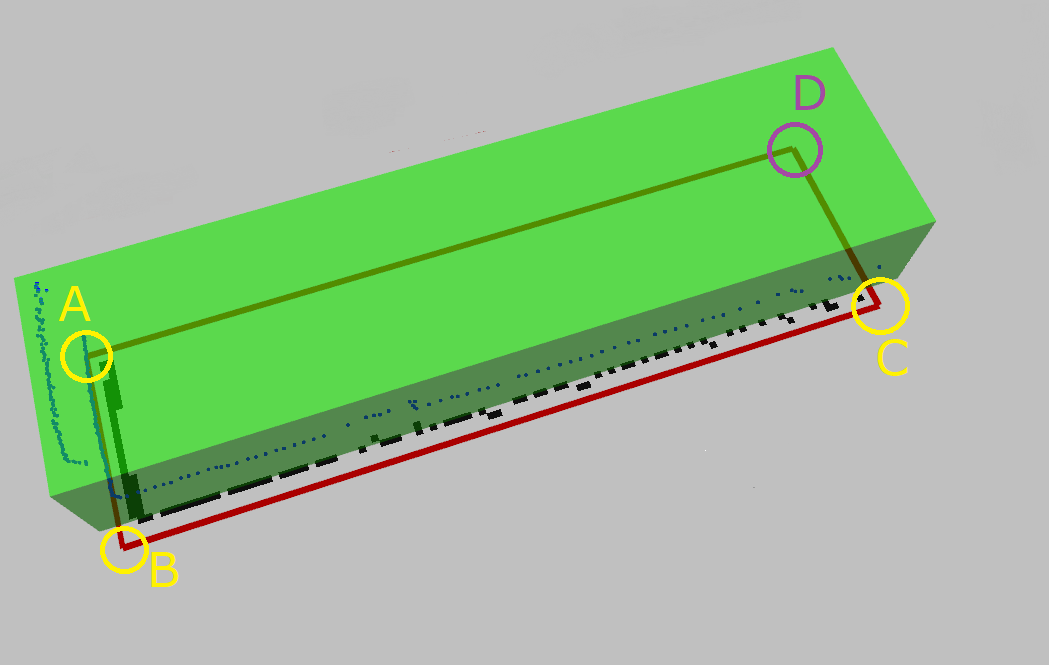}
         \caption{3D bounding box computed retrieving the 3 yellow corner (A,B,C) and extrapolating the purple one (D).}
         \label{fig:bus_full}
     \end{subfigure}
     \\
     \begin{subfigure}[b]{0.4\textwidth}
         \centering
         \includegraphics[width=\textwidth]{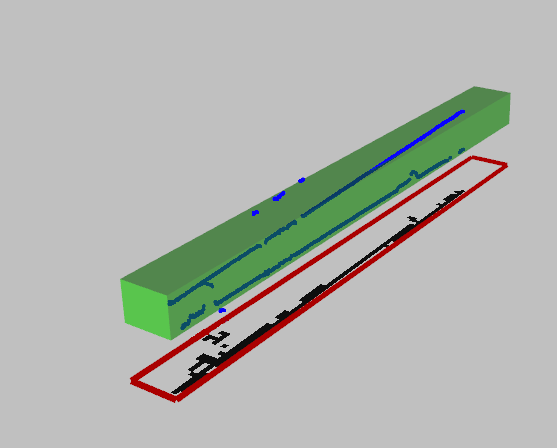}
         \caption{3D bounding box with heading computed using RANSAC, since only one side is visible.}
         \label{fig:bus_side}
     \end{subfigure}
     \hfill

        \caption{3D bounding boxes of a bus, computed few seconds apart. In the first scenario, only the side of the vehicle is visible, and the heading is computed using RANSAC. In the second scenario, both sides are visible, and the box is computed extrapolating the D point under the rectangular assumption.}
        \label{fig:8_plane_fit}
\end{figure}

The main differences are in the clustering. In the previous algorithm, we retrieved from the grid only 2D bounding boxes, which were used to extract a portion of the original pointcloud, and compute all the information on that one. In this case, since fitting a vertical plane on a single lidar layer is not feasible, we want to retrieve the heading from the 2D grid. To do so, after computing the connected components, we proceed to reconstruct a convex hull for each element. Then we differentiate two possible scenarios, displayed in Fig.~\ref{fig:8_plane_fit}. 

If both dimensions of the convex hull are above a fixed threshold, it means the obstacle is not perpendicular with respect to the lidar, and therefore two sides are visible, Fig.~\ref{fig:bus_full}. In this case, the convex hull generally has a triangular shape, since only two sides of an obstacle are visible. Accordingly, we retrieve the three vertices of the triangle and infer from those the possible fourth vertex. From those four points, it is possible to calculate the heading of the obstacle. In particular, to filter noise, we compute the relative angle of each side and average it, with the assumption of a rectangular shape for the obstacle. 

If one size is under the threshold, we assume that the obstacle is parallel or perpendicular to the lidar, Fig.\ref{fig:bus_side}. In this case, it is impossible to employ the described approach, and computing the heading based only on two points, which might be close together, would be excessively noisy. Therefore, we proceed to fit a line using RANSAC algorithm~\cite{ransac} on the occupancy grid points. The slope of this line is the heading of the obstacle. This solution is still noisier than computing the values on the two obstacle side, but it allows us to retrieve the heading even when only one side of the obstacle is visible.

To increase the accuracy, we still take the cropped area of the original pointcloud around this 2D bounding box, using the pointcloud point to compute the size and center of the box and a minimum height. Since some obstacles might be described by only one lidar plane, we are not able to return the exact height of the obstacle, but only the maximum value retrieved by the lidar. 

In this pipeline, we assume that obstacles have a rectangular shape to compute the heading. Since this requirement is not satisfied by a pedestrian, we check the convex hull size before performing the heading computation. If both dimensions are under a fixed threshold and are more similar to a pedestrian than a vehicle, this computation is not performed. We still compute the 3D box using the information from the grid and the original pointcloud, but we will not provide a heading. 

Computing the heading on a discretized grid introduces a small noise compared to the original points. To improve this, a solution might be to use the cropped projected pointcloud as input to RANSAC. After an experimental analysis where we compared the values computed using the grid and the one from the projected 2D points, we concluded that the improvements of this solution are minimal. In contrast, the increased computational power required to fit a line with RANSAC directly on the pointcloud points is substantial. For this reason, we preferred the approach previously illustrated, which is computationally less expensive and still guarantees high enough accuracy.

\section{experimental results}

\begin{figure*}[t]
     \centering
     \begin{subfigure}[b]{0.48\textwidth}
         \centering
         \includegraphics[width=\textwidth]{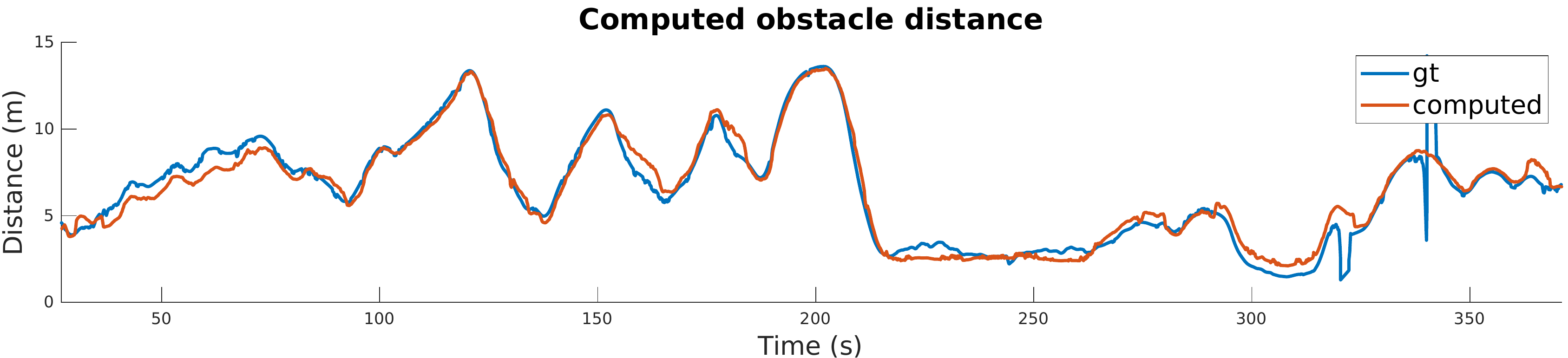}
         \caption{Computed distance between the obstacle and the ego vehicle.}
         \label{fig:16_plane_dist}
     \end{subfigure}
     \hfill
     \begin{subfigure}[b]{0.48\textwidth}
         \centering
         \includegraphics[width=\textwidth]{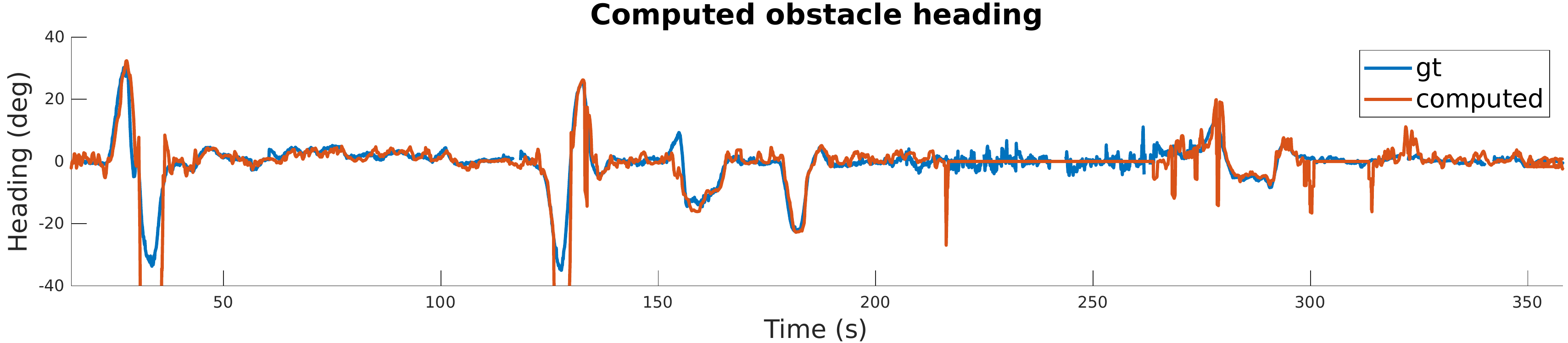}
         \caption{Computed relative heading between the obstacle and the ego vehicle.}
         \label{fig:16_plane_head}
     \end{subfigure}
     \hfill

        \caption{Computed distance and relative heading between the obstacle and the ego vehicle using the 16-plane based approach.}
        \label{fig:16_plane}
\end{figure*}

\begin{figure*}[t]
     \centering
     \begin{subfigure}[b]{0.48\textwidth}
         \centering
         \includegraphics[width=\textwidth]{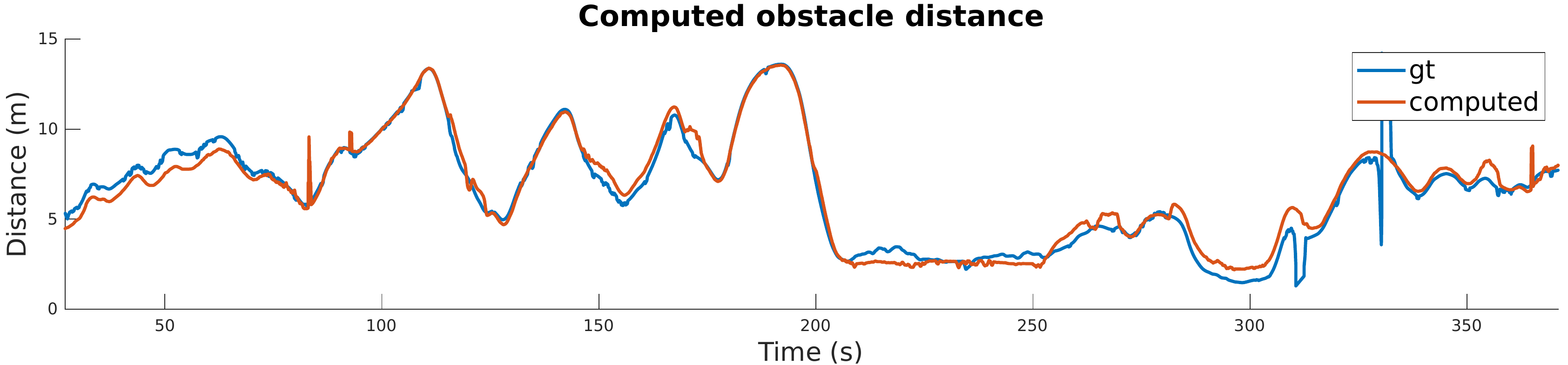}
         \caption{Computed distance between the obstacle and the ego vehicle.}
         \label{fig:8_plane_dist}
     \end{subfigure}
     \hfill
     \begin{subfigure}[b]{0.48\textwidth}
         \centering
         \includegraphics[width=\textwidth]{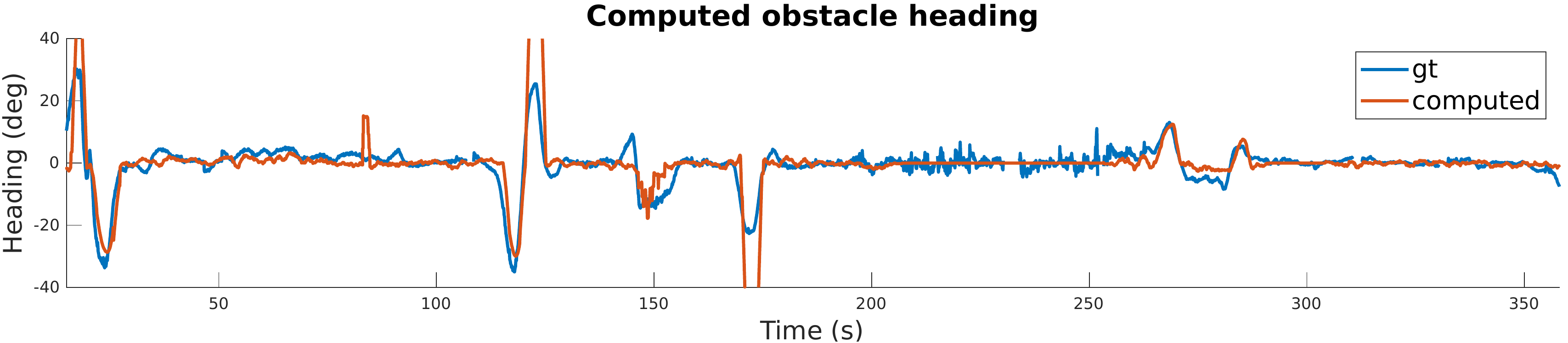}
         \caption{Computed relative heading between the obstacle and the ego vehicle.}
         \label{fig:8_plane_head}
     \end{subfigure}

        \caption{Computed distance and relative heading between the obstacle and the ego vehicle using the 8-plane based approach.}
        \label{fig:8_plane}
\end{figure*}

Both algorithms have been validated on a dataset acquired in the Monza ENI circuit. The recorded data consists of a full lap, with data from a 16 plane lidar and RTK-GPS position of the ego vehicle and one obstacle. The obstacle is a medium-sized van, which performs multiple maneuvers around the ego vehicle, changing its distance from zero up to twenty meters. From this recording, it is possible to compare the distance and heading computed by the algorithms with respect to accurate ground truth. The second approach is designed to work with lidars with low numbers of planes, therefore we employed a decimated version of the pointcloud. Moreover, the algorithm works mainly on a 2D grid, and the quality of the output is not heavily dependant on the number of planes. The only advantage provided by the higher number of planes is in the obstacles' height estimation.

Fig.~\ref{fig:16_plane} shows the results of the first approach, based on 16 planes lidar, on the recorded data. In particular, it is possible to notice how the computed distance follows the ground truth for the whole lap, Fig.~\ref{fig:16_plane_dist}, even when this value grows up to $15m$. The mean distance error of the recording is $0.7m$, which is an acceptable value for this task, also considering some section of the track where the ground truth has some small error due to GPS occlusion. Moreover, due to the particular geometry of the obstacles, the mounting point of the GPS sensor is not exactly on the back of the vehicle. Therefore this value might be slightly lower than the one computed. Nevertheless, the computed distance can be considered accurate enough for the planning algorithm to compute a trajectory and safely drive the ego vehicle. 

Fig.~\ref{fig:16_plane_head} shows instead the estimated relative heading. In this scenario, the computed values are close to the ground truth, with a mean error of $0.1\degree$. This extremely low value is partially due to the mean absolute value of the run, which is close to zero. The algorithm performs well also in some challenging sections of the track, through high curvature corners and large chicanes (i.e., $Time$ $150s$ and $Time$ $270s$). While it still follows the ground truth closely in the two really fast chicanes (i.e., $Time$ $20s$ and $Time$ $130s$), overshooting only when the angle between the ego vehicle and the obstacle is above $40\degree$.

The results of the second approach are shown in Fig.~\ref{fig:8_plane}. Also, in this case, it is possible to notice how the computed values accurately follow the ground truth. The distance error is slightly higher than the 16-planes solution, with a mean error of $0.8m$. Of particular interest is how the algorithm can still compute an accurate position when the obstacle is far from the ego vehicle (e.g., $15m$). The most significant error can be instead identified when the obstacle is close and on one side of the ego vehicle (e.g., $Time$ $280s$), where the number of points describing the obstacle is low due to the mounting position of the lidar and the proximity of the obstacle. The same issue can also be identified in the previous approach, proving that this problem is more due to physical constraints of the system than related to a specific implementation. This is also why most autonomous vehicles employ other sensors for lateral detection at close range, like sonar and single-point lasers. 

The heading in the second scenario is also noisier, with a higher mean error (while in the 16 plane case, we had $0.09\degree$ in this scenario, the error is $0.12\degree$). This value is not fully representative due to the long section with low heading, but it is possible to notice how the computed value is noisier than the first versions of the algorithm. Moreover, in the high curvature corners, the overshoot is more pronounced and, while following the ground truth closely, it is possible to see a higher error. After some analysis, we concluded that this higher error is  due to the discretization process of the grid on which values are computed. Indeed, computing these values on a discretized surface will always be less accurate than using the real 3D points of the cloud. Nevertheless, Fig.~\ref{fig:8_plane_head} shows how the algorithm is still able to compute values close to the ground truth. Therefore, while less accurate, this solution can also be employed as a source for the control algorithm.
\section{conclusions}

In this paper, we presented two algorithms for obstacle detection from sparse pointcloud. The first one is designed to work with 16-plane lidars, where deep-learning-based approaches are not suitable due to the low number of features, but have enough points to allow us to perform vertical plane fitting operations. The second one is instead developed to work with all types of sensors, since it performs most of the operations on a 2D occupancy grid. Therefore, the only missing information with a lower number of planes is the height of the obstacles. Both solutions have been validated using a custom acquired dataset, with accurate ground truth, to compare the real obstacle position and heading with the one from the algorithms. Both solutions have proved their ability to compute 3D bounding boxes with low error. The second approach is slightly less accurate due to the grid discretization process, but the error values are acceptable for control. Moreover, the solution can run in real-time on a consumer laptop without a modern GPU. 

Future works will be centered on implementing a final block of the pipeline to perform classification on the retrieved bounding box, similarly to neural network-based approaches. A second improvement will focus on tracking the state of each obstacle, in such a way, it should be possible to mitigate the bounding box noise, and filter spikes in the heading estimation.

\bibliographystyle{IEEEtran}
\bibliography{root.bbl}

\begin{thebibliography}{10}
\providecommand{\url}[1]{#1}
\csname url@rmstyle\endcsname
\providecommand{\newblock}{\relax}
\providecommand{\bibinfo}[2]{#2}
\providecommand\BIBentrySTDinterwordspacing{\spaceskip=0pt\relax}
\providecommand\BIBentryALTinterwordstretchfactor{4}
\providecommand\BIBentryALTinterwordspacing{\spaceskip=\fontdimen2\font plus
\BIBentryALTinterwordstretchfactor\fontdimen3\font minus
  \fontdimen4\font\relax}
\providecommand\BIBforeignlanguage[2]{{%
\expandafter\ifx\csname l@#1\endcsname\relax
\typeout{** WARNING: IEEEtran.bst: No hyphenation pattern has been}%
\typeout{** loaded for the language `#1'. Using the pattern for}%
\typeout{** the default language instead.}%
\else
\language=\csname l@#1\endcsname
\fi
#2}}

\bibitem{ASVADI2016299}
\BIBentryALTinterwordspacing
A.~Asvadi, C.~Premebida, P.~Peixoto, and U.~Nunes, ``3d lidar-based static and
  moving obstacle detection in driving environments: An approach based on
  voxels and multi-region ground planes,'' \emph{Robotics and Autonomous
  Systems}, vol.~83, pp. 299--311, 2016. [Online]. Available:
  \url{https://www.sciencedirect.com/science/article/pii/S0921889016300483}
\BIBentrySTDinterwordspacing

\bibitem{6083105}
J.~{Larson} and M.~{Trivedi}, ``Lidar based off-road negative obstacle
  detection and analysis,'' in \emph{2011 14th International IEEE Conference on
  Intelligent Transportation Systems (ITSC)}, 2011, pp. 192--197.

\bibitem{6122515}
J.~{Han}, D.~{Kim}, M.~{Lee}, and M.~{Sunwoo}, ``Enhanced road boundary and
  obstacle detection using a downward-looking lidar sensor,'' \emph{IEEE
  Transactions on Vehicular Technology}, vol.~61, no.~3, 2012.

\bibitem{8340798}
W.~{Zong}, C.~{Zhang}, Z.~{Wang}, J.~{Zhu}, and Q.~{Chen}, ``Architecture
  design and implementation of an autonomous vehicle,'' \emph{IEEE Access},
  vol.~6, pp. 21\,956--21\,970, 2018.

\bibitem{VANBRUMMELEN2018384}
\BIBentryALTinterwordspacing
J.~{Van Brummelen}, M.~O’Brien, D.~Gruyer, and H.~Najjaran, ``Autonomous
  vehicle perception: The technology of today and tomorrow,''
  \emph{Transportation Research Part C: Emerging Technologies}, vol.~89, pp.
  384--406, 2018. [Online]. Available:
  \url{https://www.sciencedirect.com/science/article/pii/S0968090X18302134}
\BIBentrySTDinterwordspacing

\bibitem{lang2019pointpillars}
A.~H. Lang, S.~Vora, H.~Caesar, L.~Zhou, J.~Yang, and O.~Beijbom,
  ``Pointpillars: Fast encoders for object detection from point clouds,'' in
  \emph{Proceedings of the IEEE/CVF Conference on Computer Vision and Pattern
  Recognition}, 2019, pp. 12\,697--12\,705.

\bibitem{fankhauser2016universal}
P.~Fankhauser and M.~Hutter, ``A universal grid map library: Implementation and
  use case for rough terrain navigation,'' in \emph{Robot Operating System
  (ROS)}.\hskip 1em plus 0.5em minus 0.4em\relax Springer, 2016, pp. 99--120.

\bibitem{bersani2021integrated}
M.~Bersani, S.~Mentasti, P.~Dahal, S.~Arrigoni, M.~Vignati, F.~Cheli, and
  M.~Matteucci, ``An integrated algorithm for ego-vehicle and obstacles state
  estimation for autonomous driving,'' \emph{Robotics and Autonomous Systems},
  vol. 139, p. 103662, 2021.

\bibitem{44033}
J.~Borenstein and Y.~Koren, ``Real-time obstacle avoidance for fast mobile
  robots,'' \emph{IEEE Transactions on Systems, Man, and Cybernetics}, vol.~19,
  no.~5, pp. 1179--1187, 1989.

\bibitem{grisettiimproved}
G.~Grisetti, C.~Stachniss, and W.~Burgard, ``Improved techniques for grid
  mapping,'' \emph{Robotics, IEEE Transactions on}, pp. 1--12.

\bibitem{30720}
A.~Elfes, ``Using occupancy grids for mobile robot perception and navigation,''
  \emph{Computer}, vol.~22, no.~6, pp. 46--57, 1989.

\bibitem{hornung13auro}
\BIBentryALTinterwordspacing
A.~Hornung, K.~M. Wurm, M.~Bennewitz, C.~Stachniss, and W.~Burgard,
  ``{OctoMap}: An efficient probabilistic {3D} mapping framework based on
  octrees,'' \emph{Autonomous Robots}, 2013. [Online]. Available:
  \url{http://octomap.github.com}
\BIBentrySTDinterwordspacing

\bibitem{5602377}
N.~Suganuma and T.~Matsui, ``Robust environment perception based on occupancy
  grid maps for autonomous vehicle,'' in \emph{Proceedings of SICE Annual
  Conference 2010}, 2010, pp. 2354--2357.

\bibitem{BADUE2021113816}
\BIBentryALTinterwordspacing
C.~Badue, R.~Guidolini, R.~V. Carneiro, P.~Azevedo, V.~B. Cardoso, A.~Forechi,
  L.~Jesus, R.~Berriel, T.~M. Paixão, F.~Mutz, L.~{de Paula Veronese},
  T.~Oliveira-Santos, and A.~F. {De Souza}, ``Self-driving cars: A survey,''
  \emph{Expert Systems with Applications}, vol. 165, p. 113816, 2021. [Online].
  Available:
  \url{https://www.sciencedirect.com/science/article/pii/S095741742030628X}
\BIBentrySTDinterwordspacing

\bibitem{yang2020lidar}
G.~Yang, S.~Mentasti, M.~Bersani, Y.~Wang, F.~Braghin, and F.~Cheli, ``Lidar
  point-cloud processing based on projection methods: a comparison,'' in
  \emph{2020 AEIT International Conference of Electrical and Electronic
  Technologies for Automotive}.\hskip 1em plus 0.5em minus 0.4em\relax IEEE,
  2020, pp. 1--6.

\bibitem{engelcke2017vote3deep}
M.~Engelcke, D.~Rao, D.~Z. Wang, C.~H. Tong, and I.~Posner, ``Vote3deep: Fast
  object detection in 3d point clouds using efficient convolutional neural
  networks,'' in \emph{2017 IEEE International Conference on Robotics and
  Automation (ICRA)}.\hskip 1em plus 0.5em minus 0.4em\relax IEEE, 2017.

\bibitem{zhou2018voxelnet}
Y.~Zhou and O.~Tuzel, ``Voxelnet: End-to-end learning for point cloud based 3d
  object detection,'' in \emph{Proceedings of the IEEE Conference on Computer
  Vision and Pattern Recognition}, 2018, pp. 4490--4499.

\bibitem{qi2017pointnet}
C.~R. Qi, H.~Su, K.~Mo, and L.~J. Guibas, ``Pointnet: Deep learning on point
  sets for 3d classification and segmentation,'' in \emph{Proceedings of the
  IEEE conference on computer vision and pattern recognition}, 2017, pp.
  652--660.

\bibitem{s18103337}
\BIBentryALTinterwordspacing
Y.~Yan, Y.~Mao, and B.~Li, ``Second: Sparsely embedded convolutional
  detection,'' \emph{Sensors}, vol.~18, no.~10, 2018. [Online]. Available:
  \url{https://www.mdpi.com/1424-8220/18/10/3337}
\BIBentrySTDinterwordspacing

\bibitem{zermas2017fast}
D.~Zermas, I.~Izzat, and N.~Papanikolopoulos, ``Fast segmentation of 3d point
  clouds: A paradigm on lidar data for autonomous vehicle applications,'' in
  \emph{2017 IEEE International Conference on Robotics and Automation
  (ICRA)}.\hskip 1em plus 0.5em minus 0.4em\relax IEEE, 2017, pp. 5067--5073.

\bibitem{mentasti2019multi}
S.~Mentasti and M.~Matteucci, ``Multi-layer occupancy grid mapping for
  autonomous vehicles navigation,'' in \emph{2019 AEIT International Conference
  of Electrical and Electronic Technologies for Automotive (AEIT
  AUTOMOTIVE)}.\hskip 1em plus 0.5em minus 0.4em\relax IEEE, 2019, pp. 1--6.

\bibitem{ransac}
\BIBentryALTinterwordspacing
M.~A. Fischler and R.~C. Bolles, ``Random sample consensus: A paradigm for
  model fitting with applications to image analysis and automated
  cartography,'' \emph{Commun. ACM}, vol.~24, no.~6, p. 381–395, June 1981.
  [Online]. Available: \url{https://doi.org/10.1145/358669.358692}
\BIBentrySTDinterwordspacing

\end{thebibliography}

\end{document}